\documentclass[letterpaper, 10pt]{article}
\usepackage[a4paper,top=2cm,bottom=2cm,left=2.5cm,right=2.5cm,marginparwidth=1.75cm]{geometry}

\makeatletter
\DeclareRobustCommand{\change}{%
  \@bsphack
  \leavevmode 
  \color{red} 
  \@esphack
}
\DeclareRobustCommand{\stopchange}{%
  \@bsphack
  \normalcolor
  \@esphack
}
\makeatother

\usepackage{graphics} 
\usepackage{graphicx}
\usepackage{epsfig} 
\usepackage{mathptmx} 
\usepackage{times} 
\usepackage{amsmath} 
\usepackage{amssymb}  
\usepackage[ruled,vlined]{algorithm2e}
\usepackage{booktabs}
\usepackage{tikz}
\usepackage{array}
\usetikzlibrary{arrows,positioning,fit,shapes,calc}
\usepackage{multirow}
\usepackage{multicol}
\usepackage[caption=false]{subfig}
\usepackage{xcolor}
\usepackage[export]{adjustbox}

\usepackage[inline]{enumitem}

\newcommand{\eg}{\emph{e.g.}, }
\newcommand{\ie}{\emph{i.e.}, } 
\newcommand{\etal}{\textit{et al.}, }

\title{\LARGE \bf
IBURD: Image Blending for Underwater Robotic Detection}

\author{Jungseok Hong$^1$, Sakshi Singh$^2$, and Junaed Sattar$^3$
\thanks{The authors are with the Department of Computer Science and Engineering, Minnesota Robotics Institute, University of Minnesota--Twin Cities, 100 Union St SE, Minneapolis, MN, 55455, USA
{\tt\small \{$^1$jungseok,$^2$sing0975,$^3$junaed\}@umn.edu.} This work was supported by the National Science Foundation Award IIS-2220956.
}}
\date{}

\begin{document}
\maketitle
\thispagestyle{empty}
\pagestyle{empty}

\begin{abstract}
We present an image blending pipeline, \textit{IBURD}, that creates realistic synthetic images to assist in the training of deep detectors for use on underwater autonomous vehicles (AUVs) for marine debris detection tasks. 
Specifically, IBURD generates both images of underwater debris and their pixel-level annotations, using source images of debris objects, their annotations, and target background images of marine environments. 
With Poisson editing and style transfer techniques, IBURD is even able to robustly blend transparent objects into arbitrary backgrounds and automatically adjust the style of blended images using the blurriness metric of target background images. 
These generated images of marine debris in actual underwater backgrounds address the data scarcity and data variety problems faced by deep-learned vision algorithms in challenging underwater conditions, and can enable the use of AUVs for environmental cleanup missions. 
Both quantitative and robotic evaluations of IBURD demonstrate the efficacy of the proposed approach for robotic detection of marine debris. 
\end{abstract}

\section{Introduction}
\label{sec:intro}
The ever-increasing amounts of underwater debris pose a significant threat to the aquatic ecosystem, and their volume far exceeds the collection capacity of manned missions.
Due to the scale of this problem and the risks posed to cleanup personnel, a robotic detection and removal system becomes an attractive option.
Recent developments in machine learning and computer vision have made highly accurate detections possible in many terrestrial and aerial applications (\eg medical~\cite{litjens2017survey}, agricultural~\cite{zhang2020applications}, manufacturing~\cite{ahmad2022deep}). 
However, underwater robotic detection remains difficult due to significant visual challenges: 
\begin{enumerate*}
\item absorption makes colors vary at different depths, 
\item light scattering causes images to be noisy and blurry, and 
\item light refraction distorts objects' appearances.
\end{enumerate*}
Furthermore, usable datasets for training underwater object detection are limited. 
Object detectors trained on large terrestrial datasets cannot be used as they do not represent deformations seen in marine debris, and few sufficiently large underwater datasets~\cite{fulton2019robotic, hong2020trashcan} are available. 
\begin{figure}[t]
  \centering
  \includegraphics[width=0.7\linewidth]{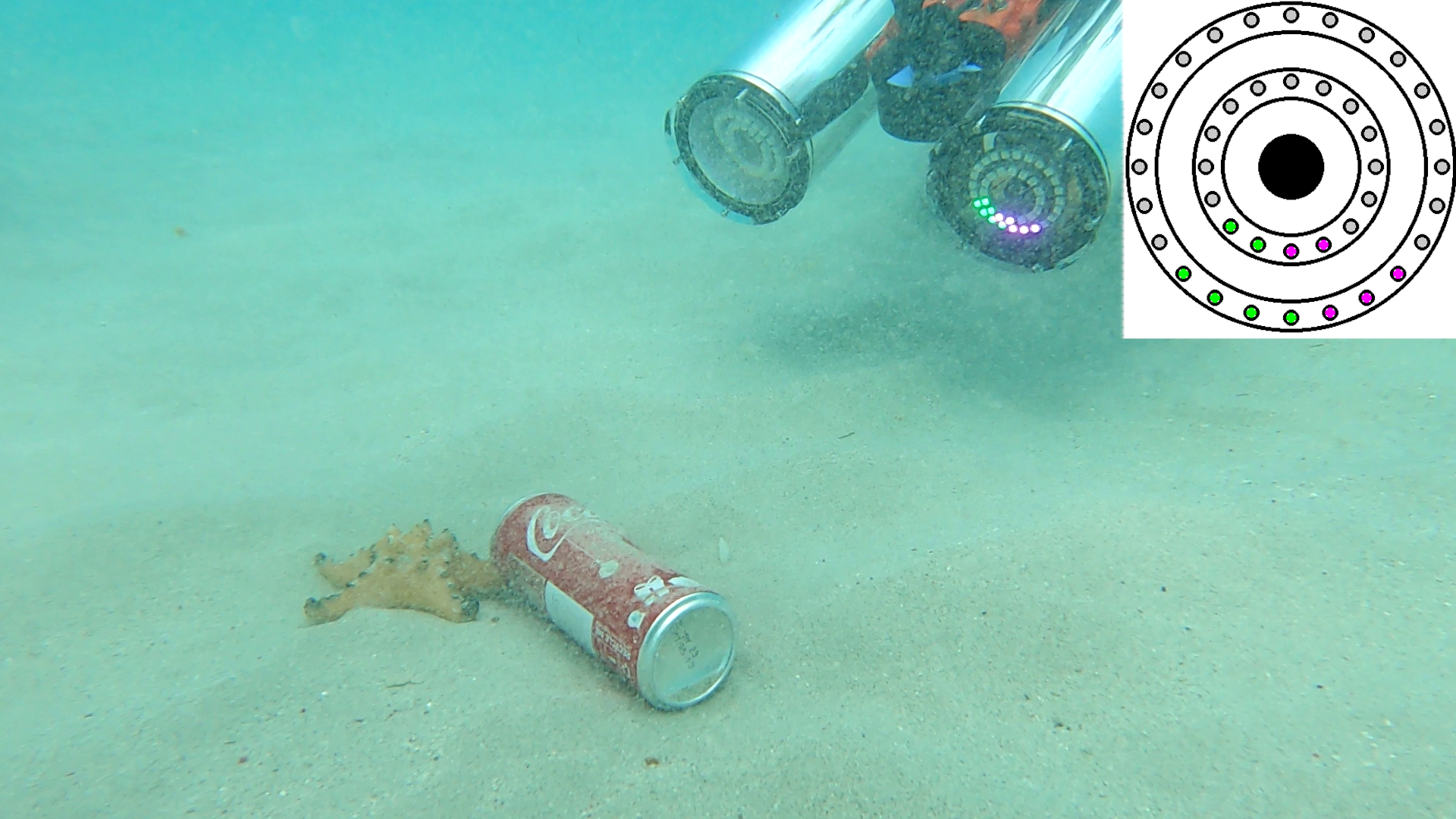}
  \caption{Demonstration of object detection on board the LoCO AUV~\cite{loco_paper_2020} in the Caribbean sea. The detection model is only trained on a synthetic dataset generated with our proposed pipeline. Magenta denotes starfish detection and green denotes can detection, where the color corresponds to the illuminated LEDs on the left ``eye''~\cite{fulton2023hreyes} of the robot. Both objects are successfully detected.}
  \label{fig:intro}
\end{figure}

Researchers have addressed marine debris detection using sonar~\cite{Toro2016Trash}, acoustic sensors~\cite{acousticdetect}, and LIDAR~\cite{ge_semi-automatic_2016}. 
Fulton \etal~\cite{fulton2019robotic} use deep learning-based underwater debris detection models and obtain higher accuracies compared to previous methods~\cite{Toro2016Trash, acousticdetect, ge_semi-automatic_2016}.
In~\cite{hong2020trashcan}, we further improve the detection performance by introducing a larger-size dataset, TrashCan, with pixel-level annotations.
Yet, building a large dataset is labor-intensive and costly. Additionally, collecting imagery of underwater debris is not just difficult and dangerous, but also infeasible, given the vastness of the underwater environment and the challenges it poses to human exploration.  
To obtain large datasets without collecting images, ~\cite{hong_generative_2020} introduces a generative method using a two-stage variational autoencoder. 
However, the images from this approach still need to be manually annotated for augmenting datasets which often becomes a bottleneck. 
Also, generative approaches have an implicit bias for the dataset they are trained on. 
We posit, and indeed discover (Sec.~\ref{sec:results}), that a detector trained on these images would not perform well in a drastically different test environment.

In this paper, we propose a framework called \textit{I}mage \textit{B}lending for \textit{U}nderwater \textit{R}obotic \textit{D}etection, or \textit{IBURD}.
This framework semi-automatically augments datasets for training object detectors. We create source object images using a text-to-image generator~\cite{ramesh2021zero} and infer their annotations with an image segmenter~\cite{kirillov2023segment}. Alternatively, it is also possible to directly use an object image and its annotation as source inputs. 
With the source images and their annotations, IBURD blends the images into target backgrounds using Poisson editing~\cite{perez_poisson_2003} and style transfer~\cite{zhang_deep_2020} techniques. 

With Poisson editing, we patch source object images with varied scales and orientations at desired locations on target background images. 
Then, style transfer matches the style of the patched objects to the target background image.
Furthermore, we update the annotations simultaneously as the orientation, scale, and location of source objects change. 
As a result, we can create realistic synthetic images with desired objects and their pixel-level annotations without collecting and labeling images manually.

We quantitatively show that augmenting a dataset with our method improves the performance of image detection and image segmentation. 
Additionally, by deploying the detector trained with the synthetic datasets on the LoCO AUV~\cite{loco_paper_2020} in open-water (sea) and confined-water (pool) environments (Fig.~\ref{fig:intro}), we show that autonomous underwater vehicles can perform object detection on mobile, low-power computing hardware in visually challenging and data-scarce environments without collecting real data prior to their deployment. 
We make the following contributions in this paper:
\begin{enumerate}
    \item We propose a novel pipeline, IBURD, to perform image blending and style transfer in series to generate realistic synthetic data semi-automatically for training object detectors.
    \item We use a novel weight adjustment approach for our loss using spatial frequency information of an image.
    \item We demonstrate that the model trained with augmented data using IBURD improves the detector's performance compared to training it with only real-world data.
    \item We evaluate the performance of the trained detector model on an AUV in pool and sea environments.
\end{enumerate}

\section{Related Work}
\label{sec:related}

Recent advances~\cite{lecun2015deep, zaidi2022survey} in deep learning have vastly improved object detection and instance segmentation results in the terrestrial domain. 
Such progress has been achieved by developing effective designs of models and training them with large datasets~\cite{lin2014microsoft, russakovsky2015imagenet} containing millions of images and corresponding labels. 
Even with such advances, detecting underwater debris still remains challenging. 
While~\cite{fulton2019robotic} presents the first deep learning based approach to detect underwater debris and outperforms previous non deep learning approaches, the accuracy is worse than general object detection tasks due to a small training dataset. 
To increase the debris detection accuracy,~\cite{hong2020trashcan} proposes a larger dataset, TrashCan, which has both bounding box and pixel-level annotations for object detection and instance segmentation along with baseline results using Mask R-CNN~\cite{he2017mask} and Faster R-CNN~\cite{ren2015faster}. 
However, increasing the dataset size to improve debris detection accuracy further is not scalable due to debris data scarcity and labeling costs. 
To overcome the data scarcity issue,~\cite{hong_generative_2020} proposes a generative method, augmenting the existing dataset with synthetic underwater debris images. 
While the method can create realistic synthetic images, it still requires additional labeling efforts to be used for training detectors. 

Style transfer~\cite{singh_neural_2021,jing_neural_2020} is an approach for changing the appearance of one image based on the visual style of another. 
\cite{rodriguez_domain_2019, yu_sc-uda_2022} use this to improve detection in images taken from various domains (\eg different light conditions and image clarity). 
They aim to account for low-level texture changes in images by updating them to have the same style throughout the data. 
\cite{kadish_improving_2021} also attempts to improve detection using style transfer, by having the detector learn high-level features (\eg object shape) instead of low-level features (\eg the texture of paintings). 
\cite{amirkhani_enhancing_2021} uses style transfer to simulate various types of noise that may be present in real-world data. 
\cite{lin_gan-based_2021,liu_lane_2020} use style transfer to imitate varying light conditions. 
Style transfer has been applied beyond RGB images; \eg\cite{cygert_style_2019} converts RGB images from COCO dataset~\cite{lin2014microsoft} to thermal images and uses them to train a thermal image detector. 
While style transfer works well in augmenting the appearance of an image, it does not add new objects to our data.

Unlike style transfer, image blending based methods allow placing new objects anywhere on target background images. 
\cite{perez_poisson_2003} introduces Poisson editing using Laplacian information to smooth the boundary between the image patches and target images. 
\cite{wu_gp-gan_2019} uses a GAN-based approach for image blending, producing realistic images; however, it requires image pairs of empty backgrounds and objects placed in the backgrounds to train, limiting its use when the source data is limited. 
\cite{georgakis_synthesizing_2017} modifies~\cite{perez_poisson_2003} to find spaces within a given image plane to blend an object. 
However, detectors trained with their synthetic data show degraded performance on real data due to the style discrepancy between the blended objects and backgrounds in the dataset.
\cite{zhang_training_2022} uses a harmonization blending approach to create new data for aerial search and rescue, but it does not blend the boundary of target objects. 

\cite{zhang_deep_2020} presents a two-stage deep network-based approach to blend an image patch onto a background. Unlike~\cite{wu_gp-gan_2019} their approach does not need additional training data to generate blended images.
\begin{figure}  
    \centering
    \scalebox{0.75}{\input{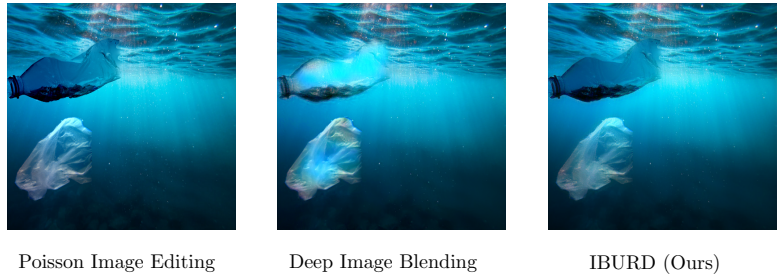}}
    \caption{Comparison of generated images using three approaches:  Poisson image editing~\cite{perez_poisson_2003}, Deep image blending~\cite{zhang_deep_2020} and our method, IBURD. In our approach, we can successfully prevent over-stylization of the blended objects.}  
    \label{fig:compare}
    \vspace{-4mm}
\end{figure}

They use the proposed method mainly for artistic purposes and it struggles with blending transparent source images onto background images, as seen in Fig.~\ref{fig:compare}. 
The method is only tested with $20$ images and takes approximately $4$ minutes to blend one object in an image of size $512\times512$ pixels.

Our proposed approach, IBURD, allows us to place source images at various locations and scales in target background images with relevant bounding box and pixel-level annotations within $50$ seconds, which is $5$ times faster than~\cite{zhang_deep_2020}. 
Our method addresses blending transparent objects using Poisson editing, a situation that previous methods fail to cover.
Additionally, IBURD deals with object distortion due to excessive style transfer using Fast Fourier Transform (FFT)~\cite{liu_image_2008} based weight adjustment for loss.

\section{Methodology}
\label{sec:methods}
\begin{figure*}[b!]
    \centering
    \scalebox{0.9}{\input{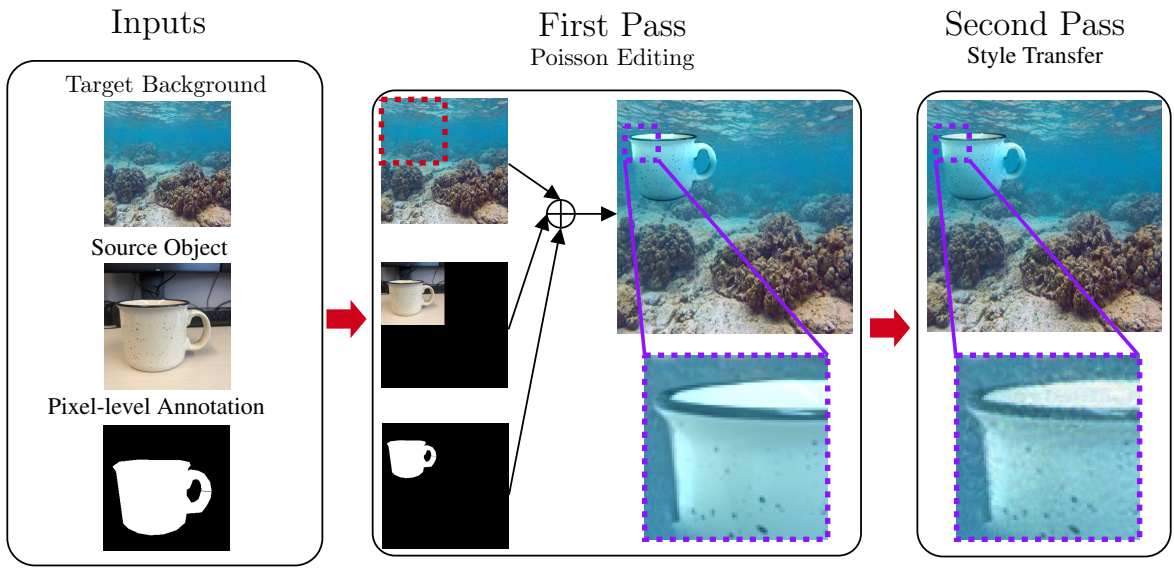}}
    \caption{IBURD pipeline: The source image with its pixel-level annotation and background image are fed as inputs. The first pass resizes and rotates the source object and selects a location for blending it. Poisson editing smooths the boundary between the object and the background. The second pass changes the style to produce the final image. In the zoomed-in region of the object, the style difference between the appearance of the first and second pass image is visible.}  
    \label{fig:pipeline}
\end{figure*}

The IBURD (Fig.~\ref{fig:pipeline}) pipeline takes an object image, its pixel-level annotation, and the target background image as inputs. IBURD then uses a two-pass process similar to~\cite{zhang_deep_2020}, consisting of Poisson editing and style transfer. Poisson editing blends an object onto a background, and style transfer changes the appearance of the object. 

While it is possible to capture source images from real objects, it is difficult to collect source images of marine life which mainly exist underwater (\eg starfish, crab, etc.). To address this, we use a text-to-image generator, Dall-E~\cite{ramesh2021zero}, to generate an object from prompts (\eg perished plastic container with white background, deformed soda can on white background). Next, we use the Segment Anything Model (SAM)~\cite{kirillov2023segment} to segment the object.
From the user provided prompt, this pre-processing step generates both an image and its annotation, which makes this step semi-automatic. Once this pair is available, we run the first pass.

In the \textit{first pass}, we randomly resize and rotate the object image along with its annotation. 
To place the object image in the background, we split the given background image into a grid and place the object image in a randomly selected cell within the grid. Details on grid size are provided in Sec.~\ref{subsec:data_gen}. We place one object in each grid cell. This is to avoid overlapping of blended objects. 
We then use Poisson image editing~\cite{perez_poisson_2003} for blending, eliminating drastic boundary gradient changes between the object and the background.

Next, we feed the output from the first pass and the background image to the \textit{second pass}, which reconstructs the first pass output to reflect the background style (\ie underwater appearance). For this, we use the total loss (Eq.~\ref{eq:total}) consisting of three different loss functions: \textit{style loss} (Eq.~\ref{style}~\cite{gatys_image_2016, zhang_deep_2020}), \textit{content loss} (Eq.~\ref{content}~\cite{gatys_image_2016, zhang_deep_2020}), and \textit{total variation loss} (Eq.~\ref{tv}~\cite{mahendran_understanding_2015, zhang_deep_2020}). 

\begin{equation}
    \label{style}
    Loss_{style} = \sum_{l=1}^{L} \frac{\beta_l}{2N_l^2} \sum_{i=1}^{N_l} \sum_{k=1}^{M_l}(G_l[I_{r}]-G_l[I_{b}])_{ik}^2
\end{equation}
\begin{equation}
    \label{content}
    Loss_{content} = \frac{\alpha_l}{2N_LM_L} \sum_{i=1}^{N_L} \sum_{k=1}^{M_L}(F_L[I_{r}]-F_L[I_{fp}])_{ik}^2
\end{equation}
\begin{equation}
    \label{tv}
    Loss_{tv} = \sum_{m=1}^{H} \sum_{n=1}^{W} |I_{m+1,n} - I_{m,n}| + |I_{m,n+1} - I_{m,n}|
\end{equation}

\begin{equation}
    \label{eq:total}
    Loss_{total} = \lambda Loss_{style} + \mu Loss_{content} + \nu Loss_{tv}
\end{equation}

\textit{Style loss} compares the background and the reconstructed image, and \textit{Content loss} reflects the differences between the blended image from the first pass and the reconstructed image.
Using style and content loss together prevents the object from being over-stylized.

In Eq.~\ref{style} and \ref{content}, $I_r$ is the \textit{reconstructed image}, $I_b$ is the \textit{background image}, $I_{fp}$ is the \textit{output from the first pass}, $L$ is the \textit{number of convolution layers}, $N_l$ is the \textit{number of channels in activation}, $M_l$ is the \textit{number of flattened activation values} used by each channel, $F_l$ is the \textit{activation matrix} computed using network $F$ at layer $l$, $G_l = F_lF_l^T$ is the \textit{Gram matrix}, $\alpha_l$ and $\beta_l$ are the \textit{respective weights}.

We compute the style and content loss using VGG-$16$~\cite{liu_very_2015} network’s layers. Specifically, style loss uses layers relu$1$\_$2$, relu$2$\_$2$, relu$3$\_$3$, and relu$4$\_$3$ ($L=4$). Content loss uses layer relu$2$\_$2$ ($L=1$). We use pretrained weights for VGG-$16$~\cite{liu_very_2015}.
Unlike~\cite{wu_gp-gan_2019}, we do not need additional training data for image generation which makes this method favorable for small-dataset scenarios.
In addition, \textit{total variation loss} preserves the low-level features in an image.  In Eq.~\ref{tv}, $I$ is the \textit{reconstructed image}. $I_{m,n}$ denotes the entry on $m^{th}$ row and $n^{th}$ column.
In Eq.~\ref{eq:total}, $\lambda$ represents the \textit{style loss weight}, $\mu$ represents \textit{content loss weight}, and $\nu$ represents \textit{total variation loss weight}.

\begin{figure*}
  \centering
  \subfloat[Generated image samples before and after the second pass. With the increase in weight, the distortion of object boundary increases.\label{weights}]{
    \scalebox{0.9}{\input{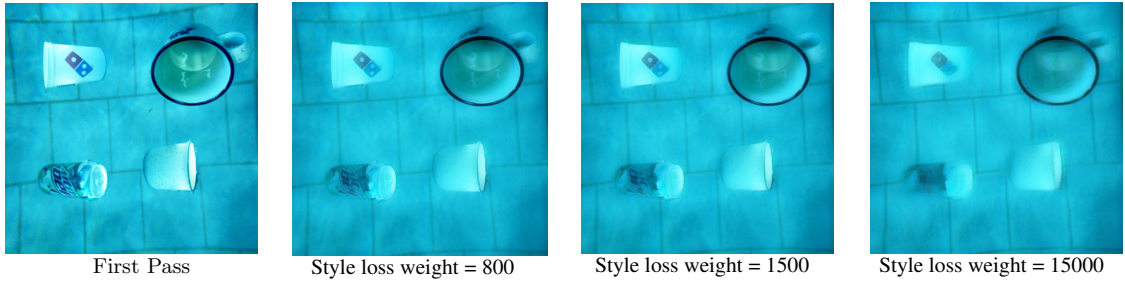}}
  }
  \hspace{25pt}
  \subfloat[Sample images used in the survey to quantify blurriness. The scores obtained by the survey are shown under respective images. A lower score signifies clear image and a higher score corresponds to increased blurriness.\label{fig:blur}]{
    \scalebox{0.9}{\input{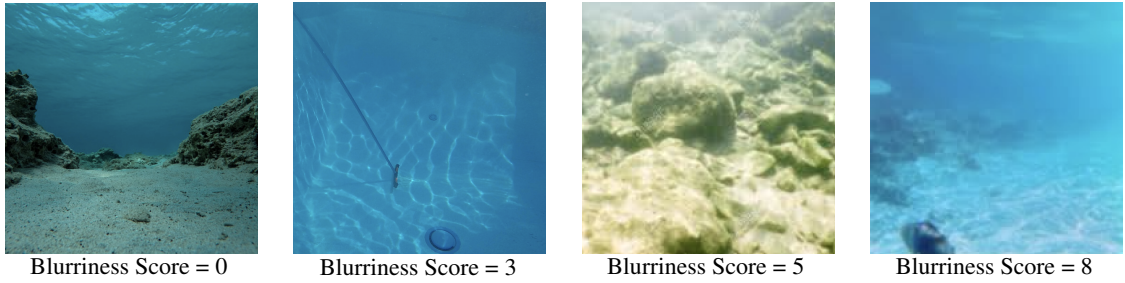}}
  }
  \caption{Object content is lost when blending objects in a blurry background. (a) shows a gradual content loss on increasing style loss weight. (b) shows examples of blurry backgrounds.}
  \label{fig:main}

\end{figure*}

Using typical style transfer approaches (\eg\cite{zhang_deep_2020,wu_gp-gan_2019}) developed for artistic purposes, the total loss (Eq.~\ref{eq:total}) distorts the object boundaries when higher style loss weight (\eg $\lambda=15000$ in Fig.~\ref{weights}) is used on a blurry background image.
However, images for training detectors need to maintain the object's content while reflecting the style of the background. 
In other words, we must regulate the style loss weight to account for background quality while not losing the object shape completely. To achieve this, we use the FFT to compute spatial frequency as a measure of image blurriness~\cite{liu_image_2008}. Using FFT on the background image, we compute a mean frequency value, which acts as an indicator of blur -- the lower the FFT value, the blurrier the image.
Note that while other measures (\cite{laplacian_blur}, \cite{wavlet_blur}) of blurriness are usable, we choose the FFT out of runtime and efficiency considerations, and it has shown good performance in that regard.

\begin{figure}
\end{figure}

\begin{table}
\caption{Style loss weight ($\lambda$) assigned based on the FFT mean value of background image.}
\label{survey_weight}
\begin{center}
\begin{tabular}{ccc}
\toprule
    FFT mean value & Avg. survey score & Style loss weight $\lambda$\\
    \midrule
     mean$>=40$ & 0 & $30000$ \\
     $40>$mean$>=10$ & 1.4 & $15000$ \\
     $10>$mean$>=0$ & 4.5 & $1500$ \\
     $0>$mean & 7.5 & $800$ \\
\bottomrule
\end{tabular}
\end{center}
\end{table}

We conduct a survey with $10$ participants who were asked to rate $10$ images on a scale of $0$ to $10$ ($0$ being clear and $10$ being extremely blurry). Examples of the survey images with their blurriness scores are shown in Fig.~\ref{fig:blur}. We use this survey to validate the correlation between an FFT measure and the human perception of blurriness in an image.

We verify the correlation between the survey and FFT mean scores and map the ranges of FFT mean values to four style loss weight values via empirical evaluations (Table~\ref{survey_weight}). A clear image has a lower blurriness score from the survey and a higher value from the FFT. For such an image, we can transfer the style of the background without losing the object content, and hence, we use a higher style loss weight. The coefficient value ($\lambda$) is dynamically assigned using the FFT mean value range mentioned in Table~\ref{survey_weight}.

Along with the blending process, the pipeline generates annotation information on the newly reconstructed image by translating and transforming the original annotation coordinates based on the blending location. We collect this information in COCO format, suitable for training common object detectors.

\section{Experiments}
\label{sec:experiments}

We measure IBURD's efficacy with quantitative evaluations using $10$ classes from the TrashCan~\cite{hong2020trashcan} dataset. We also perform robotic experiments with the LoCO AUV~\cite{loco_paper_2020} to evaluate our approach on visually similar but unseen environments (\eg pool and sea).
In both cases, we start by collecting some source object images and background images that are representative of the environment we selected to evaluate a detector. 
We then use IBURD to blend the object images into the selected backgrounds and train a detector using this synthetic data generated from the pipeline.

\begin{table}[t]

\caption{Class distribution of source object image used for quantitative evaluation.}
\label{data_dist_quant}
\begin{center}
\begin{tabular}{ccc}
\hline
\toprule
    Class name & No. of source images using Dall-E & No. of source images from TrashCan\\
    \midrule
     animal\_starfish & $3$ & $4$ \\
     trash\_bag & $3$ & $22$\\
     animal\_shell & $3$ & $2$\\
     animal\_crab & $2$ & $5$ \\
     trash\_pipe & $3$ & $4$\\
     trash\_bottle & $3$ &$5$\\
     trash\_snack\_wrapper & $3$ & $3$\\
     trash\_can & $3$ & $3$ \\
     trash\_cup & $3$ & $2$\\
     trash\_container & $3$ & $11$\\
     
\bottomrule
\end{tabular}
\end{center}
\end{table}

\subsection{Data Collection}
\subsubsection{Quantitative Experiments}
To evaluate a detector's performance on the TrashCan dataset, we select $10$ background images within the dataset. 
We choose images that are representative of the TrashCan dataset and contain as few objects in each image as possible. 
We collect source images both semi-automatically ($29$ images from Dall-E) and manually ($68$ images from TrashCan). 
Both Dall-E and TrashCan images have the same $10$ object classes  (Table~\ref{data_dist_quant}).
Unlike the source images from Dall-E, the distribution of the ones collected from TrashCan is non-uniform across the classes since TrashCan has more data from common debris items (\eg bottle) compared to other classes (\eg starfish).

\subsubsection{Robotic Experiments}
We collect $47$ source images containing $7$ classes of objects commonly found in marine debris (Table~\ref{data_dist}). We have a smaller number of classes compared to the quantitative experiments since not all $10$ classes of objects can be easily placed underwater (\eg pipe and crab).
We then manually annotate the images, providing class labels, bounding boxes, and segmentation information.
We use $2$ types of background images for blending: sea and pool. 
We collect $7$ images captured at different locations from the pool used for experiments and add $3$ pool images from online sources to add more variety. 
For the sea background, we use $10$ background images sourced from the Internet.

\subsection{Image Blending and Data Generation} 
\label{subsec:data_gen}
\subsubsection{Quantitative experiments}
\label{subsec:blend_quant}
To generate a diverse dataset with $512\times512$ pixel-size images, we use $4$ different rotations ($0^{\circ}$, $90^{\circ}$, $180^{\circ}$, $270^{\circ}$) for the source object images. 
For Dall-E generated images, we use $4$ different source image sizes,
 ($96\times96$, $128\times128$, $192\times192$, $256\times256$ pixels). 
With source objects from TrashCan, we only use $2$ source image sizes ($192\times192$, $256\times256$ pixels).
This is because TrashCan has a mixture of close-up and long-range views of objects, and using the smaller scales might produce blended objects that are not visible to the human eye, making them impossible to annotate. Close-up version of objects generated by Dall-E enables the use of smaller-scale images.

For both cases, the image is split into a $2\times2$ grid to avoid overlap with previously blended objects. We determine the object location by randomly selecting one section within the grid on the background image plane. Using the grid we blend up to $4$ objects in the same background. This way we create $3$ different datasets: 1) TrashCan training data with $2$k images generated using Dall-E source objects (\textit{T+D$2$k}), 2) TrashCan training data with $10$k images generated using Dall-E source objects (\textit{T+D$10$k}), and 3) TrashCan training data with $10$k images generated using source objects from TrashCan (\textit{T+T$10$k}). 
For these $3$ cases, we also generate images using just Poisson image editing (\ie first pass only) to train a detector on Poisson-blended data alone. This makes it possible to assess the effect of adding the second pass in the IBURD pipeline on detector performance compared to the same detector trained on basic Poisson-blended data. For both blending techniques (Poisson Image Editing and IBURD), the images used are the same in terms of objects, size, orientation, and location (\eg Fig.~\ref{fig:compare}). We use IBURD generated data for training and evaluate the performance using real-world validation data from TrashCan for all cases.

\begin{table}[b!]
\caption{Class distribution of source object image used for synthetic pool and ocean data creation.}
\label{data_dist}
\begin{center}
\begin{tabular}{cc}
\hline
\toprule
    Class name & No. of source images\\
    \midrule
     Bag & $7$ \\
     Bottle & $6$ \\
     Glass Bottle & $5$ \\
     Cup & $8$ \\
     Mug & $10$ \\
     Can & $9$ \\
     Starfish & $2$ \\
\bottomrule
\end{tabular}
\end{center}
\end{table}
\subsubsection{Robotic Experiments}
Similar to the case of quantitative experiments, we generate datasets with $4$ rotations and $4$ sizes for the $47$ source object images and create images with up to $4$ objects blended in the same background.
We determine the object location by randomly selecting one section within the grid on the background image plane. 
For multi-object blending, we divide the background ($512\times512$ pixels) into $2\times2$ grid considering the maximum size of source images (\ie $256\times256$ pixels). 
For the single object case, the image plane is divided based on the current source image size (\ie $4\times4$ grid for $96\times96$ and $128\times128$ pixel-size images, $2\times2$ grid for $192\times192$ and $256\times256$ pixel-size images). 
For the sea backgrounds, we create $1,880$ images with $1$ object, $2,209$ images with $2$ objects, $3,008$ images with $3$ objects, and $4,096$ images with $4$ objects making a total of $11,193$ images for training. 
Similarly, in the case of pool background, we create $1,880$ images with $1$ object, $2,209$ images with $2$ objects, $2,396$ images with $3$ objects, and $1,731$ images with $4$ objects, giving a total of $8,216$ images for training. 
We generate a smaller dataset for pool images since it is a controlled environment and has limited variety in the type of backgrounds that can occur in real-world images.

\begin{figure}[t]

  \centering
  \includegraphics[width=.3\linewidth]{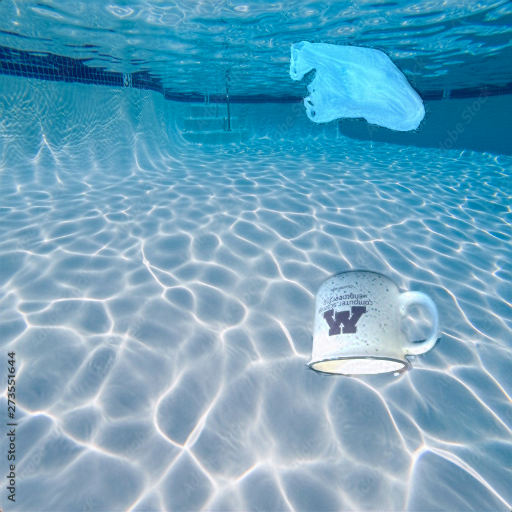}
  \hspace{2mm}
  \includegraphics[width=.3\linewidth]{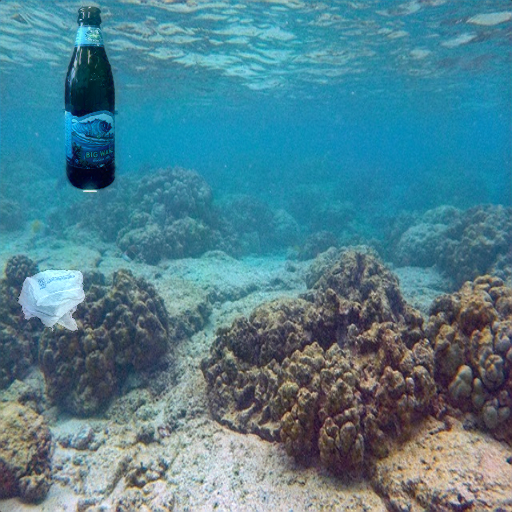}\\
  \vspace{2mm}
  \includegraphics[width=.3\linewidth]{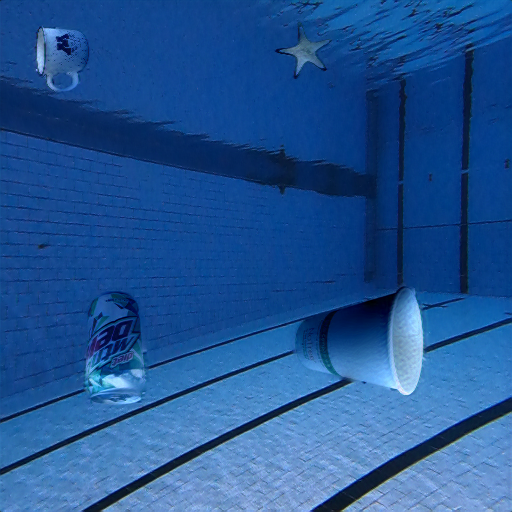}
  \hspace{2mm}
  \includegraphics[width=.3\linewidth]{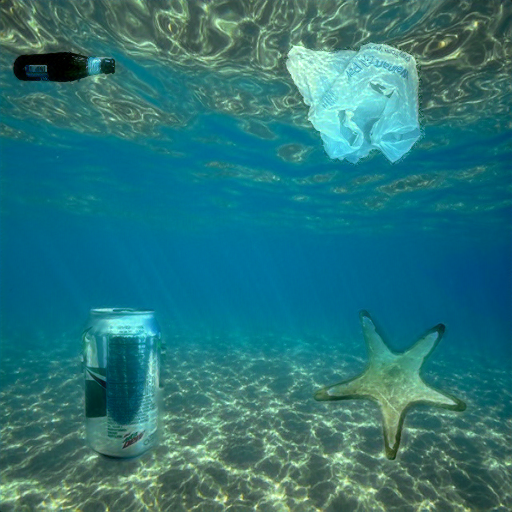}
  \caption{Sample images generated from IBURD. The first column shows images blended on pool backgrounds. The second column contains objects blended on ocean backgrounds.}
  \label{blend_example}
\end{figure}

In both experiments, the background image and the final blended image are of size $512\times512$. We use $100$ iterations for style transfer during the second pass. We empirically fix the number of iterations. Similar to ~\cite{zhang_deep_2020}, we use a $L-BFG$ solver to optimize the total loss, set the content loss weight $\mu$ to $1$, and choose the total variation loss weight $\nu$ to be $10^{-6}$. In our method, the style loss weight $\lambda$ is based on image blurriness (Table~\ref{survey_weight}). Some examples of generated images are shown in Fig.~\ref{blend_example}.

\subsection{Object Detection and Instance Segmentation}
To evaluate the efficacy of our framework, we first select YOLACT~\cite{bolya2019yolact} as an instance segmentation model based on its inference speed and performance with pretrained weights. 
We choose ResNet50-FPN as a backbone of YOLACT to obtain a reasonable inference time on the low-power mobile GPU (see Sec.~\ref{sec:robot_setup}) on the LoCO AUV.
We train the model with $4$ different datasets for quantitative experiments (Sec.~\ref{subsec:blend_quant}): \textit{T+D$2$k}, \textit{T+D$10$k}, \textit{T+T$10$k}, and original TrashCan data. For robotic experiments, we train the model with $2$ synthetic datasets: pool and sea data. The model is trained on an NVIDIA GeForce RTX 2080 Ti. 

\subsection{Robot Setup}\label{sec:robot_setup}
We use the LoCO AUV to run the detector model with trained weights on its mobile GPU (\ie NVIDIA Jetson TX2). 
We utilize image frames from the right camera of the LoCO AUV to make inferences. 
We adopt an LED lighting indicator system~\cite{fulton2023hreyes} 
installed on the LoCO AUV's left camera to visually examine the performance of different network weights during deployments. 
The $40$ LEDs in the system are split into $7$ groups, where each group denotes a specific class with a unique color (as shown in Fig.~\ref{fig:images}).

\section{Results}
\label{sec:results}

Our method creates an effective dataset for training robust visual debris detectors rather than solely improving the appearance of images for aesthetic purposes. 
The desired properties of generated images are a smooth transition border between the object and background images and the preservation of the object's content.
Our pipeline satisfies these properties and removes the background in the source object image completely from transparent (\eg plastic bottles) and translucent objects (\eg glass bottles and plastic bags). 
Additionally, the objects are not heavily distorted even when other objects exist in the background (\eg coral reefs in Fig.~\ref{fig:pipeline}). Note that while FFT-based style weights do prevent distortion, an extremely blurry background could still cause the object to lose its content information.
IBURD achieves $5$ times faster runtime compared to the recent work, Deep Image Blending method~\cite{zhang_deep_2020}. Our method can generate an image with one object in $25$ seconds. The runtime increases as more objects are blended in the background, reaching up to $50$ seconds for $4$ objects.

\subsection{Quantitative Evaluation}
\begin{table}[t]
    \caption{Detector performance comparison: synthetic data generated with IBURD and generated with Poisson editing only.}
    \centering
    \begin{tabular}{cccc}
    \toprule
         Dataset & Type & mAP (IBURD) & mAP (Poisson Editing)  \\ \midrule
         T+T10k &box& 46.59 & 23.68\\ 
         T+T10k &mask& 40.17& 22.33\\ 
         T+D2k& box & 49.25 & 33.78 \\ 
         T+D2k& mask & 42.58 & 26.92  \\ 
         T+D10k& box & 49.95 & 32.59  \\ 
         T+D10k& mask & 41.77 & 24.42  \\  \bottomrule

    \end{tabular}
    \label{tab:ablation}
    
\end{table}
We train the detector on three datasets for comparing IBURD and Poisson Blending: 1) \textit{T+T$10$k}, 2) \textit{T+D$2$k}, and 3) \textit{T+D$10$k}. Adding the second pass in the IBURD pipeline improves the performance of the detector in all three cases compared to Poisson blending only, as shown in Table \ref{tab:ablation}. The \textit{T+D$2$k} dataset performs the best in the case of Poisson editing. This leads us to conclude that the detector performance deteriorates as we add more data without considering the style.

We also compare the above three cases with our baseline \ie a detector trained on real images from Trashcan (\textit{T}), and the results are shown in Table \ref{tab:map_results}. Each row represents a detector trained on different datasets (as shown in column 1), with the first two rows representing the baseline cases. All evaluations, however, are performed only using real data from the TrashCan dataset.  
Even after augmenting \textit{T} with synthetically generated $10$k images using TrashCan objects, the performance of the detector does not show much improvement in the \textit{T+T$10$k} case. 
\begin{table}[b!]
    \caption{Quantitative evaluation results for each dataset and its detection type: T (TrashCan), T+T$10$k (T+$10$k generated images with the TrashCan objects), T+D$2$k (T+$2$k generated images from Dall-E), T+D$10$k (T+$10$k generated images from Dall-E).}
    \centering
    \begin{tabular}{cccccc}
    \toprule
         Dataset & Type & AP & AP$_{50}$ & AP$_{70}$ & AP$_{90}$  \\ \midrule
         T (baseline)&box & 46.22 & 72.01 & 59.92 &	6.59 \\ 
         T (baseline)&mask & 41.33 & 70.11 & 50.26 & 12.05 \\ 
         T+T10k &box& 46.59 &70.93&59.51 & 11.85 \\ 
         T+T10k &mask& 40.17& 64.63 &51.05 & 12.97 \\ 
         T+D2k& box & 49.25 & 72.34 & 63.39 & 14.52  \\ 
         T+D2k& mask & \textbf{42.58} & \textbf{71.53} & 51.58 & 11.92  \\ 
         T+D10k& box & \textbf{49.95} & \textbf{73.35} & \textbf{64.14} & \textbf{17.12}  \\ 
         T+D10k& mask & 41.77 & 65.55 & \textbf{52.79} & \textbf{14.29}  \\  \bottomrule

    \end{tabular}
    \label{tab:map_results}
\end{table}
Additionally, augmenting data with new objects using Dall-E improves the detector performance (\textit{T+T$2$k} and \textit{T+T$10$k}). However, there is no significant difference in performance from increasing the augmented dataset size from $2$k to $10$k using the same objects.
This leads us to believe that even if it is possible to generate a large number of images using different backgrounds, sizes, etc., the detector performs better when introducing data with more significant information change.
While IBURD can provide a way to create any number of images depending on provided inputs, it is up to the user to decide how to choose the generated data distribution and what objects or information to include.
\subsection{Robotic Evaluation}
\begin{figure}
\setlength{\lineskip}{0pt}
\centering
\setlength\tabcolsep{1.pt}
\renewcommand{\arraystretch}{0.5}
  \begin{tabular}{c@{\extracolsep{0.1cm}}c@{\extracolsep{0.1cm}}c}
  \includegraphics[trim=0 60 0 58 , clip, width=0.3\linewidth]{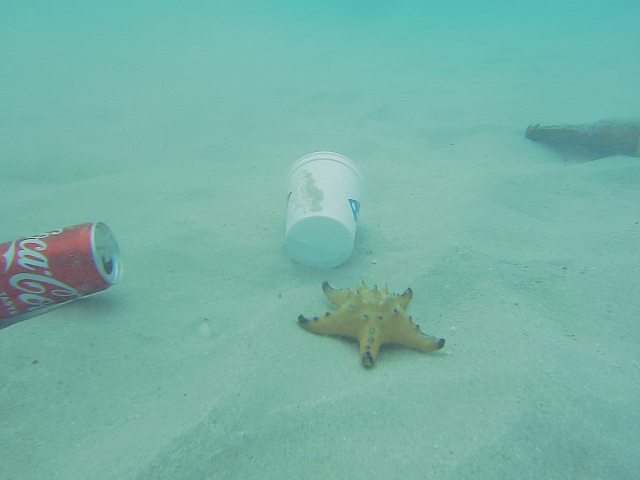} &
         \includegraphics[trim=0 60 0 58 , clip, width=0.3\linewidth]{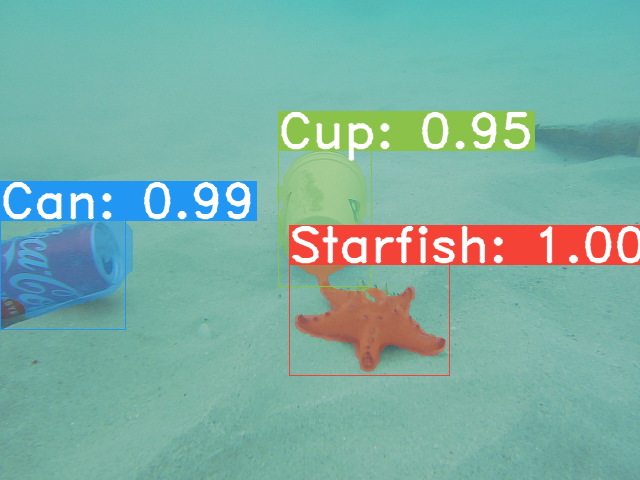} &
           \includegraphics[width=0.3\linewidth]{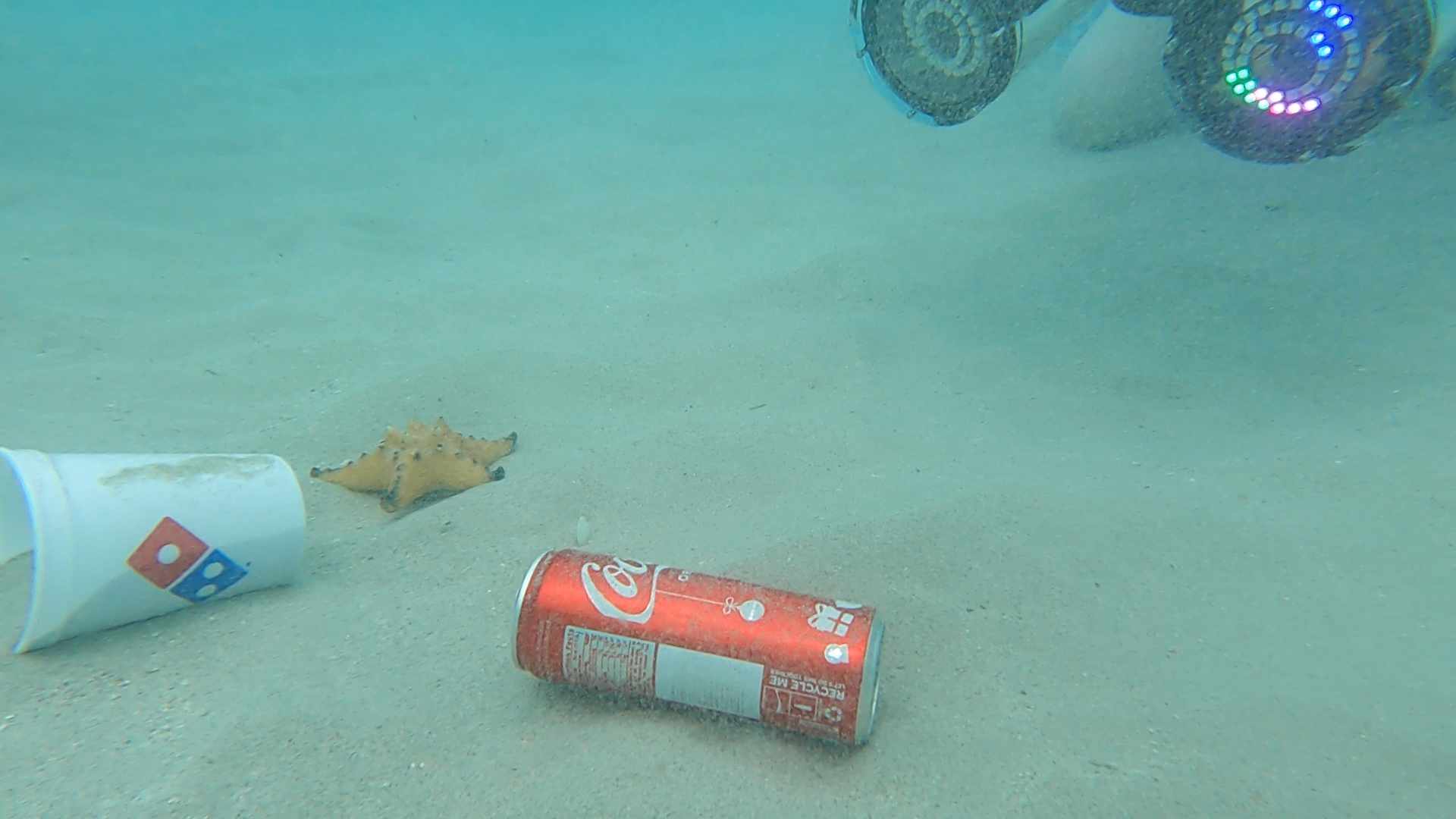}\\
 \includegraphics[trim=0 118 0 0 , clip, width=0.3\linewidth]{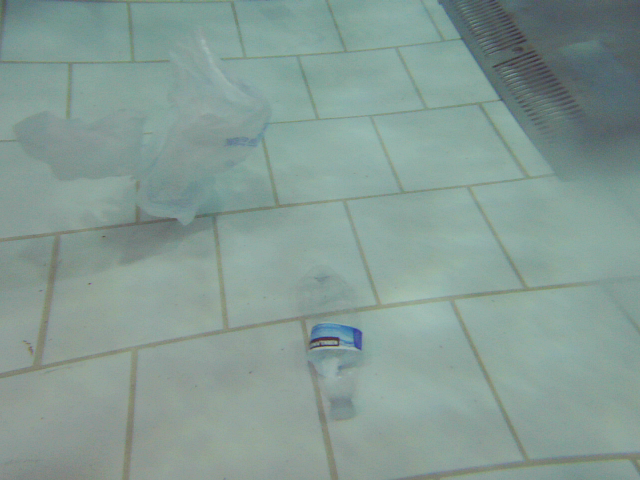}
&
     \includegraphics[trim=0 118 0 0 , clip, width=0.3\linewidth]{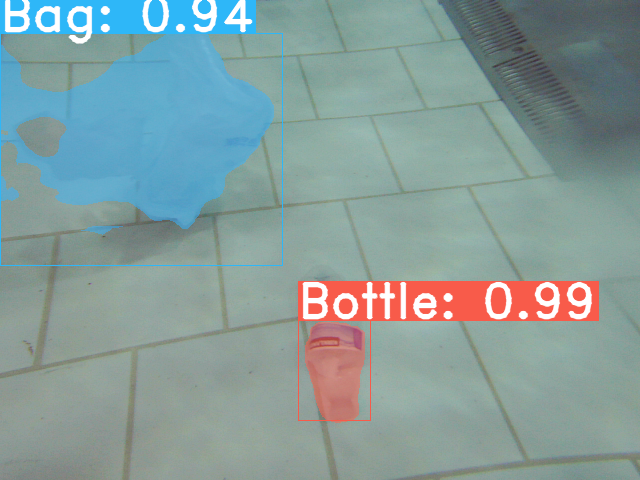} &

     \includegraphics[width=0.3\linewidth]{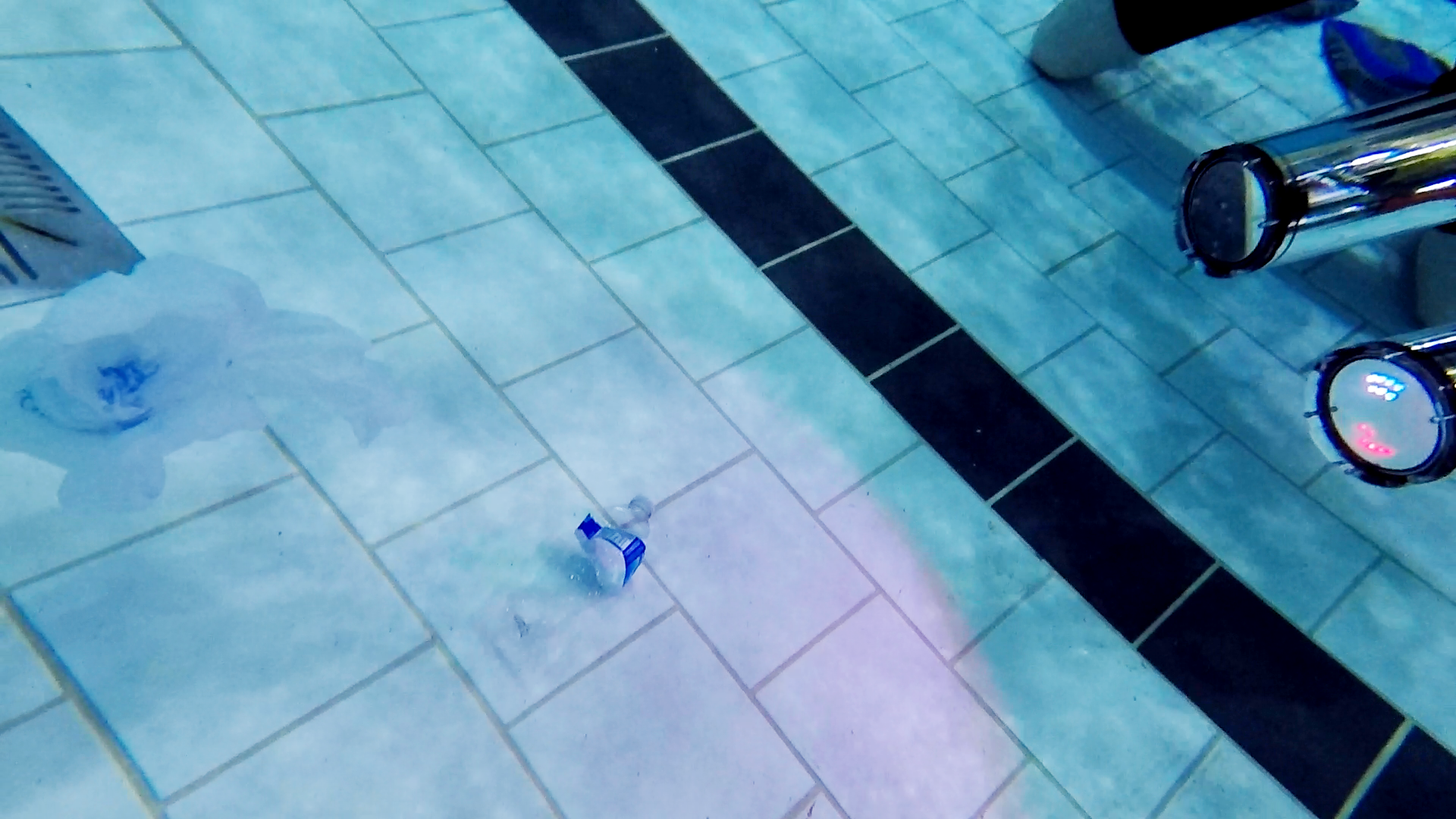}
  \\
    
    \footnotesize{(a) Raw input images} & \footnotesize{(b) Detection results} & \footnotesize{(c) Detection response using LEDs} \\

  \end{tabular}
  \caption{The first row consists of real images from the Caribbean Sea. The second row contains real images from a pool. These images are used as input for testing the detector network, which is trained on images generated using IBURD. (a) shows the raw images from the perspective of the camera on the LoCO AUV, (b) visualizes the segmentation mask and bounding box predicted by the detector. The network detects the objects correctly, with high confidence scores, and (c) shows the LED response in real-time. In the ocean image (first row) in (c) cup, starfish, and can are visible with their respective LED colors blue, magenta, and green. In the pool image (second row) in (c) bag and bottle are present with their LED indicating their class colors, white and red.} 
\label{fig:images}
\end{figure}
\begin{figure}
\setlength{\lineskip}{0pt}
\centering
\setlength\tabcolsep{1.pt}
\renewcommand{\arraystretch}{0.5}
  \begin{tabular}{c@{\extracolsep{0.05cm}}c@{\extracolsep{0.1cm}}c@{\extracolsep{0.1cm}}c@{\extracolsep{0.1cm}} c}
  & & $w_{Sea}$ & $w_{Pool}$ & $w_{TrashCan}$ \\ \\
  \raisebox{-3.\normalbaselineskip}[0pt][0pt]{{\rotatebox[origin=c]{90}{\footnotesize{Sample from}}}} &
  \raisebox{2.\normalbaselineskip}[0pt][0pt]{{\rotatebox[origin=c]{90}{\footnotesize{Sea}}}} &
    \includegraphics[width=0.25\textwidth, cfbox=blue 1pt 0.3pt]{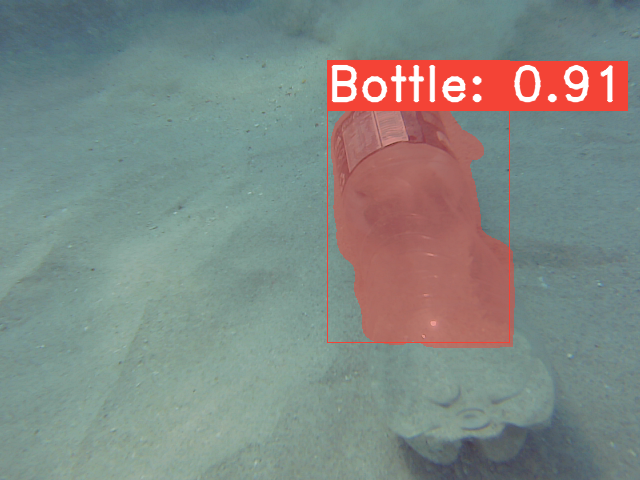} &
    \includegraphics[width=0.25\textwidth]{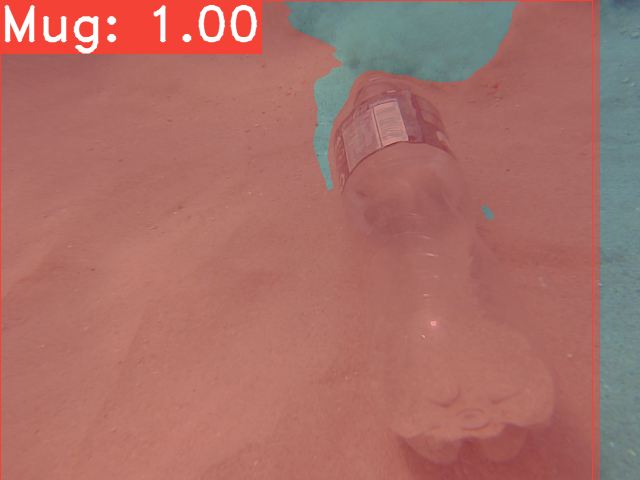} &
    \includegraphics[width=0.25\textwidth]{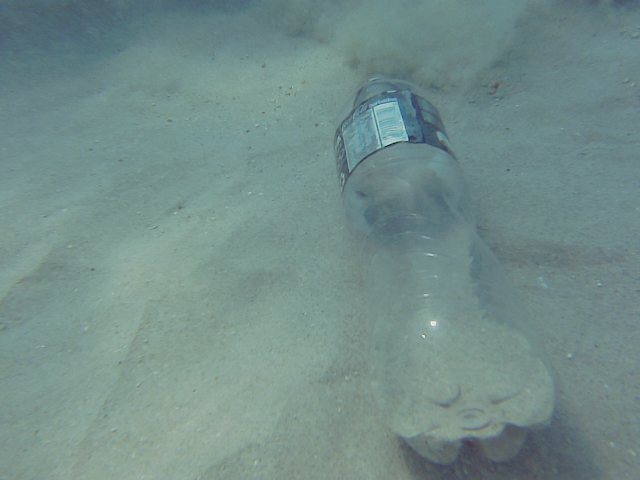} \\
      &\raisebox{2.\normalbaselineskip}[0pt][0pt]{{\rotatebox[origin=c]{90}{\footnotesize{Pool}}}}&   
    \includegraphics[trim=-0cm 0 0 -0cm, width=.24\textwidth]{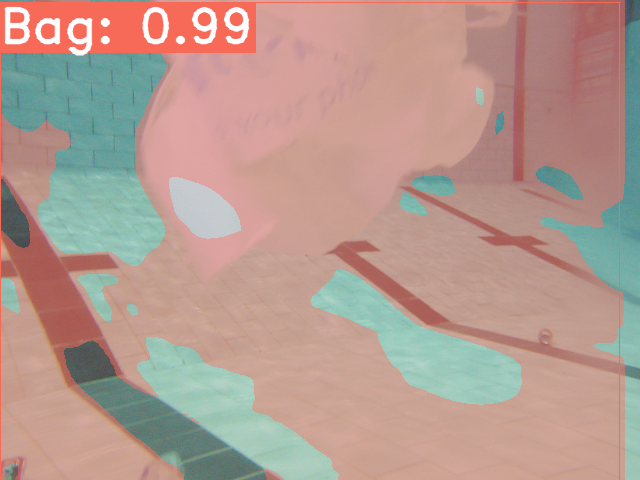} &
    \includegraphics[width=.25\textwidth, cfbox=blue 1pt 0.3pt]{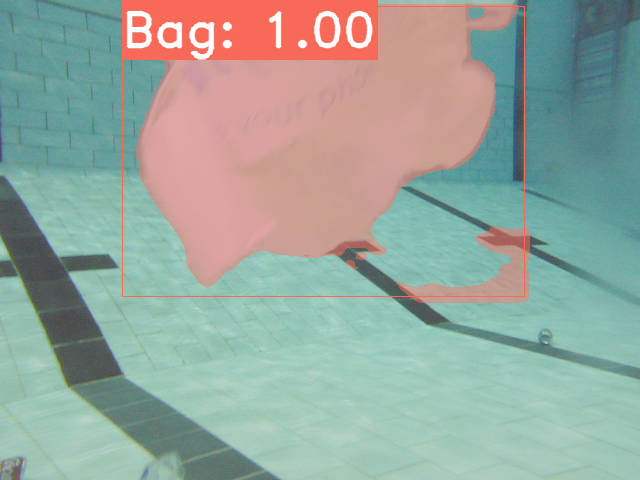} &
    \includegraphics[width=.25\textwidth]{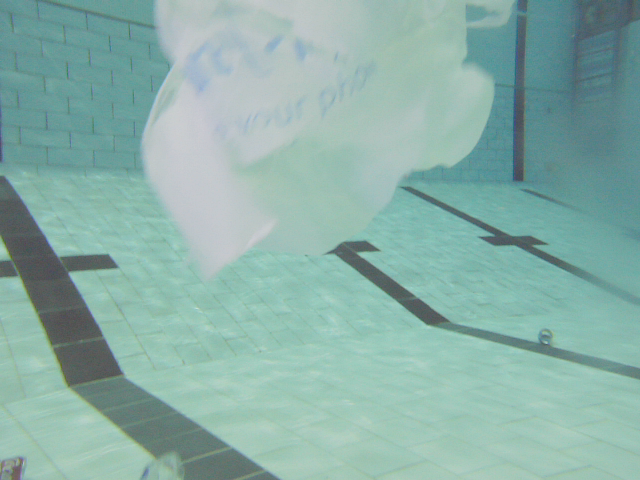}   \\
      &\raisebox{2.\normalbaselineskip}[0pt][0pt]{{\rotatebox[origin=c]{90}{\footnotesize{TrashCan}}}}&
    \includegraphics[width=.25\textwidth]{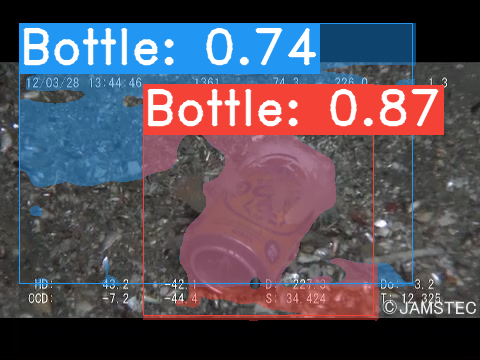} &
    \includegraphics[width=.25\textwidth]{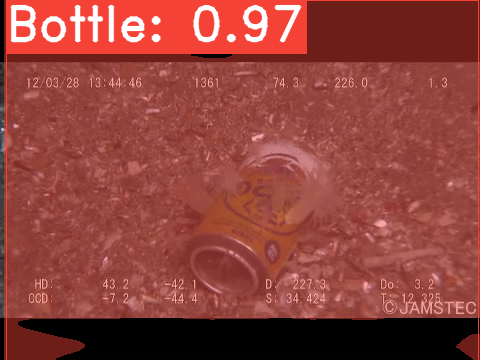} &
    \includegraphics[width=.25\textwidth, cfbox=blue 1pt 0.3pt]{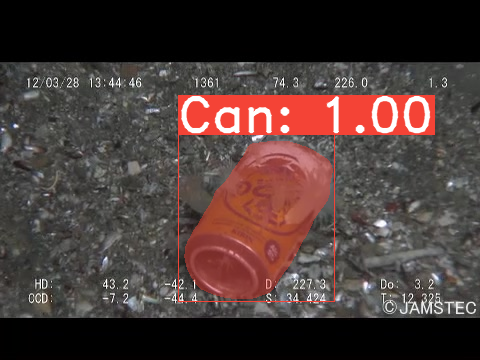} \\
  \end{tabular}
  \caption{Robotic evaluation on real-world images: Images in each row are samples from three different sources (\ie Sea, Pool, and TrashCan). Columns correspond to detector weights trained on sea synthetic ($w_{Sea}$), pool synthetic ($w_{Pool}$), and TrashCan ($w_{TrashCan}$) datasets. For each weight, the performance is evaluated on all three sample images. The detector performs best when the sample images are from a similar or the same environment as the training dataset.}
  \label{fig:sample}
\end{figure}
We train the detector on synthetically generated datasets and deploy it on the LoCO AUV. We conduct experiments in pool and sea environments as explained in Sec.~\ref{sec:experiments}. 
The detector performs at $1$-$3$ frames/second (FPS) on a NVIDIA Jetson TX2, and we monitor detector output through the LED indicator. The detector successfully infers object classes, which are in the training datasets, during pool and sea deployments (Fig.~\ref{fig:images}). 
We also evaluate the performance of detectors trained with three different datasets (\ie synthetic sea, synthetic pool, and TrashCan) using images from three different environments (\ie sea, pool, and TrashCan), in total nine cases (Fig.~\ref{fig:sample}).
When the detector is trained and tested with images from the same environments, objects are inferred correctly. Real images from the sea, pool, and TrashCan are tested on the detector with weights $w_{Sea}$, $w_{Pool}$, and $w_{TrashCan}$, respectively.
\textit{However, if the network weight and sample image pairs are from different environments, the detector shows degraded performance or completely fails to detect objects.} 
This demonstrates the necessity of using relevant datasets for target environments. 
In our experiments, we show IBURD can generate realistic synthetic datasets for pool and sea environments without using any real images of objects in the target environment. 
We also demonstrate that the detectors trained with synthetic datasets which have similar visual features to a target environment perform better than ones trained with publicly available datasets if the target environment is known a priori. The accompanying video contains additional examples of generated data and a demonstration of the on-board robot experiments

\subsection{Limitations}
The performance of a network is heavily dependent on the kind of training data used. While we do provide a pipeline to generate data, manual effort is needed to make sure the variety and the style of data match the validation data. Additionally, we do not consider overlapping objects.

\section{Conclusions}
This paper presents a two-pass image blending pipeline, \textit{IBURD}, to generate synthetic images with a given background using source objects and their annotations which can be obtained semi-automatically for underwater debris detection. 
Our approach provides pixel-level annotations for each synthetic image, eliminating the labor-intensive dataset annotation process. 
IBURD can blend transparent source object images without creating artificial borders between the object and background. 
Also, our blurriness score-based blending method can dynamically synthesize source and target background images, resulting in more realistic synthetic images than previous methods.
The results demonstrate that our pipeline enables training of object detection and instance segmentation networks suitable for the data-scarce underwater debris detection problem. 

We plan to extend the pipeline in multiple directions. 
We are looking to refine our approach for scenarios where objects overlap each other or are partially occluded to address more complex scenes.
We are also developing an approach to decide the desired locations of source objects in target background images by considering the correlation between the semantic information of the background and objects. 
Lastly, we intend to improve the blurriness evaluation algorithm by enabling selective evaluation of regions of interest for partially blurry target background images. 





\bibliographystyle{IEEEtran}
\bibliography{ref}
\end{document}